\documentclass{llncs}
\usepackage{amsmath}
\usepackage{amsfonts}
\usepackage{algorithm}
\usepackage{algorithmic}
\usepackage{mathrsfs}
\usepackage{paralist}
\usepackage{epsfig}
\usepackage{subfigure}

\def \Reals {{\mathbb{R}}}

\def \FT {{\mathfrak{T}}}

\def \FR {{\mathfrak{R}}}

\def \CA {{\mathcal{A}}}

\def \CM {{\mathcal{M}}}

\def \CS {{\mathcal{S}}}

\def \BS {{\mathbb{S}}}

\def\argmax{\mathop{\rm arg\,max}}

\spnewtheorem{assumption}{Assumption}{\bfseries}{\itshape}

\begin{document}
\frontmatter
\pagestyle{headings}  % switches on printing of running heads
\mainmatter
\title{Nearly optimal exploration-exploitation decision thresholds}
\author{Christos Dimitrakakis
\thanks{Thanks to M. Keller and R. Chavarriaga, for comments and interesting discussions. This work has received financial support from the Swiss NSF under the MULTI project
(2000-068231.021/1) and from IDIAP. This is updated version better discusses earlier work and places this paper in a proper context.}
}
\authorrunning{C. Dimitrakakis}   % abbreviated author list (for running head)
\institute{IDIAP Research Institute, 4 Rue de Simplon, Martigny CH 1920, Switzerland \email{dimitrak@idiap.ch}}
\maketitle

\begin{abstract}
  While in general trading off exploration and exploitation in
  reinforcement learning is hard, under some formulations relatively
  simple solutions exist.  In this paper, we first derive upper bounds
  to for the utility of selecting different actions in the multi-armed
  bandit setting. Unlike the common statistical upper confidence
  bounds, these explicitly link the planning horizon, uncertainty and
  the need for exploration explicit. The resulting algorithm can be
  seen as a generalisation of the classical Thompson sampling
  algorithm. We experimentally test these algorithms, as well as
  $\epsilon$-greedy and the value of perfect information heuristics.
  Finally, we also introduce the idea of bagging for reinforcement
  learning. By employing a version of online bootstrapping, we can
  efficiently sample from an approximate posterior distribution.
  \end{abstract}

\section{Introduction}
\label{sec:Introduction}
In reinforcement learning, the dilemma between selecting actions to
maximise the expected return according to the current world model and
to improve the world model such as to {\em potentially} be able to
achieve a higher expected return is referred to as the {\em
exploration-exploitation trade-off}. This has been the subject of much
interest before, one of the earliest developments being the theory of
sequential sampling in statistics, as developed by
\cite{Wald:SequentialAnalysis}.  This dealt mostly with making
sequential decisions for accepting one among a set of particular
hypothesis, with a view towards applying it to jointly decide the
termination of an experiment and the acceptance of a hypothesis.  A
more general overview of sequential decision problems from a Bayesian
viewpoint is offered in
\cite{Degroot:OptimalStatisticalDecisions}.

The optimal, but intractable, Bayesian solution for bandit problems
was given in~\cite{Bellman:1956}, while recently tight bounds on the
sample complexity of exploration have been found
\cite{jmlr:Mannor:SampleComplexity}. An approximation to the full
Bayesian case for the general reinforcement learning problem is given
in \cite{dearden98bayesian}, while an alternative technique based on
eliminating actions which are confidently estimated as low-value is
given in \cite{jmlr:Even-Dar:ActionElimination}.

As will be seen in the next section, the intuitive concept of trading
exploration and exploitation is a natural consequence of the {\em
  definition} of the problem of reinforcement learning. Our first
contribution is to see show in Sec.~\ref{sec:optimal_bandit} that
there is a simple threshold for switching from exploratory to greedy
behaviour in bandit problems.  This threshold is found to depend on
the effective reward horizon of the optimal policy and on our current
belief distribution of the expected rewards of each action.  A sketch
of the extension to MDPs is presented in
Sec.~\ref{sec:optimal_reinforcement_learning}, while
Sec.~\ref{sec:optimistic_evaluation} derives an upper bound on the
value of exploration to derive practical algorithms.
Section~\ref{sec:online-bootstr-reinf} introduces the use of online
bootstrapping for reinforcement learning, in order to approximate
sampling for posterior distributions.  Finally, the resulting
algorithms are then illustrated experimentally in
Sec. \ref{sec:optimal_bandit_experiments}.  We conclude with a
discussion on the relations with other methods.
 
\section{Exploration Versus Exploitation}
\label{sec:exploration_vs_exploitation}
Let us assume a standard multi-armed bandit setting, where a reward
distribution $p(r_{t+1}|a_t)$ is conditioned on actions in $a_t \in
\CA$, with $r_t \in \Reals$.  The aim is to discover a policy $\pi =
\{P(a_t = i) | i \in \CA\}$ for
selecting actions such that $E[r_{t+1}|\pi]$ is maximised.  It follows
that the optimal gambler, or oracle, for this problem would constitute
a policy which always chooses $i \in \CA$ such that $E[r_{t+1}|a_t=i]
\geq E[r_{t+1}|a_t=j]$ for all $j \in \CA$.  Given the conditional
expectations, implementing the oracle is trivial.  However this tells
us little about the optimal way to select actions when the
expectations are unknown. As it turns out, the optimal action
selection mechanism will depend upon the problem formulation.  We
initially consider the two simplest cases in order to illustrate that
the exploration/exploitation tradeoff is and should be viewed in terms
of problem and model definition.

In the first problem formulation the objective is to discover a
parameterized probabilistic policy $\pi = \big\{P(a_t|\theta_t) ~\big|~ a_t \in
\CA\big\}$, with parameters $\theta_t$, for selecting actions such that
$E[r_{t+1}|\pi]$ is maximised.  If we consider a model whose
parameters are the set of estimates $\theta_t = \big\{q_i =
\hat{E}_t[r_{t+1}|a_t=i] ~\big|~ i \in \CA\big\}$, then the optimal choice is to select $a_t$
for which the estimated expected value of the reward is highest,
because according to our current belief any other choice will
necessarily lead to a lower expectation.  Thus, stating the bandit
problem in this way does not allow the exploration of seemingly lower,
but potentially higher value actions and it results in a {\em greedy}
policy.

In the second formulation, we wish to minimise the discrepancy between
our estimate $q_i$ and the true expectation. This could be written as
the following minimisation problem:
\[
        \sum_{i \in \CA} E\big[\|r_{t+1} - q_i\|^2 ~\big|~ a_t=i\big].
\]
For point estimates of the expected reward, this requires sampling
{\em uniformly} from all actions and thus represents a purely
exploratory policy.  If the problem is stated as simply minimising the
discrepancy asymptotically, then uniformity is not required and it is
only necessary to sample from all actions infinitely often.  This
condition holds when $P(a_t=i) > 0 ~ \forall i
\in \CA, ~ t > 0$ and can be satisfied by mixing the optimal policies
for the two formulations, with a probability $\epsilon$ of using the
uniform action selection and a probability $1-\epsilon$ of using the
greedy action selection.  This results in the well-known
$\epsilon$-greedy policy (see for example \cite{Sutton+Barto:1998}),
with the parameter $\epsilon \in [0,1]$ used to control exploration.

This formulation of the exploration-exploitation problem, though
leading to an intuitive result, does not lead to an obvious way to
optimally select actions.  In the following section we shall consider
bandit problems for which the functional to be maximised is
\[
        E\bigg[\sum_{k=0}^N g(k) r_{t+k+1}\bigg | \pi \bigg],
                \quad g(k) \in [0,1], ~  N \geq 0,
\]
with $\sum_{k=0}^\infty g(k) < \infty$.  In this formulation of the
problem we are not only interested in maximising the expected reward
at the next time step, but in the subsequent $N$ steps, with the
$g(\cdot)$ function providing another convenient way to weigh our
preference among short and long-term rewards.  Intuitively it is
expected that the optimal policy for this problem will be different
depending on how long-term are the rewards that we are interested in.
As will be shown later, by lengthening the effective reward horizon
through manipulation of $g$ and $N$, i.e. by changing the definition
of the problem that we wish to solve, the exploration bias is
increased automatically.

\section{Optimal Exploration Threshold for Bandit Problems}
\label{sec:optimal_bandit}
We want to know when it is a better decision to take action $i$ rather
than some other action $j$, with $i,j \in \CA$, given that we have
estimates $q_i, q_j$ for $E[r_{t+1}|a_t=i]$ and $E[r_{t+1}|a_t=j]$
respectively\footnote{For bandit problems with states in a state space
$\CS$, similar arguments can be made by considering $i, j \in \CS \times
\CA$.}. We shall attempt to see under which conditions it is better to
take an action different than the one whose expected reward is
greatest. For this we shall need the following assumption:
\begin{assumption}[Expected rewards are bounded from below]
\label{ass:bounded_rewards}
There exists $b \in \Reals$ such that
\begin{equation}
\label{eq:bounded_rewards}
        E[r_{t+1}|a_t=i] \geq b \quad \forall~i \in \CA,
\end{equation}
\end{assumption}
The above assumption is necessary for imposing a lower bound on the
expected return of exploratory actions: no matter what action is
taken, we are guaranteed that $E[r_t] > b$. Without this condition,
exploratory actions would be too risky to be taken at all.

Given two possible actions to take, where one action is currently
estimated to have a lower expected reward than the other, then it
might be worthwhile to pursue the lower-valued action if the following
conditions are true: \begin{inparaenum}[(a)] \item there is a degree
of uncertainty such that the lower-valued action can potentially be
better than the higher-valued one, \item we are interested in
maximising more than just the expectation of the next reward, but the
expectation of a weighted sum of future rewards,  \item we will be
able to accurately determine whether one action is better than the
other quickly enough, so that not a lot of resources will be wasted in
exploration\end{inparaenum}.

We now start viewing $q_i$ as random variables for which we hold
belief distributions $p(q_i)$, with $\bar{q_i} = E[q_i] =
\hat{E}[r_{t+1}|a_t=i]$.  The problem can be defined as deciding when
action $i$, is better than taking action $j$, under the condition that
doing so allows us to determine whether $q_i > q_j + \delta$ with high
probability after $T \geq 1$ exploratory actions. For this reason we
will need the following bound on the expected return of exploration.
\begin{lemma}[Exploration bound]
\label{lem:exploration_bound}
For any return of the form $R_t = \sum_{k=0}^N g(k)r_{t+k+1}$, with
$g(k) \geq 0$, assuming \eqref{eq:bounded_rewards} holds, the expected
return of taking action $i$ for $T$ time-steps and following a greedy
policy thereafter, when $\bar{q_i} > \bar{q_j}$, is bounded below by
\begin{multline}
U(i, j, T, \delta, b) = \sum_{k=T}^N g(k) \big((\bar{q_j}+\delta) P(q_i > q_j + \delta) + \bar{q_j}
P(q_i \leq q_j + \delta)\big)\\
+ \sum_{k=0}^{T-1} g(k) 
\big((\bar{q_j}+\delta)  P(q_i > q_j+\delta) + b P(q_i \leq q_j+\delta)\big)
\label{eq:exploration_bound}
\end{multline}
for some $\delta > 0$.
\end{lemma}
This follows immediately from Assumption \ref{ass:bounded_rewards}.
The greedy behaviour supposes we are following a policy where we
continue to perform $i$ if we know that $P(q_i > q_j + \delta)
\approx 1$ after $T$ steps and switch back to $j$ otherwise.

Without loss of generality, in the sequel we will assume that $b=0$
(If expected rewards are bounded by some $b \neq 0$, we can always
subtract $b$ from all rewards and obtain the same).  For further
convenience, we set $p_i = P(q_i \geq q_j + \delta)$. Then we may
write that we must take action $i$ if the expected return of simply
taking action $j$ is smaller than the expected return of taking action
$i$ for $T$ steps and then behaving greedily, i.e. if the following
holds:
\begin{align}
\sum_{k=0}^N g(k) \bar{q_j}
	&< \sum_{k=T}^N g(k) \big((\bar{q_j}+\delta) p_i + \bar{q_j} (1-p_i) \big) + \sum_{k=0}^{T-1} g(k) (\bar{q_j}+\delta) p_i\\
%%\sum_{k=0}^{T-1} g(k) \big( \bar{q_j} - (\bar{q_j} + \delta) p_i \big)
%%	&< \sum_{k=T}^N g(k)  \big((\bar{q_j}+\delta) p_i + \bar{q_j} (1-p_i) - \bar{q_j} \big)\\
%%\sum_{k=0}^{T-1} g(k) \big( \bar{q_j} - (\bar{q_j} + \delta) p_i \big)
%%	&< \sum_{k=T}^N g(k)  \big((\bar{q_j}+\delta) p_i - \bar{q_j} p_i \big)\\
\sum_{k=0}^{T-1} g(k) \big( \bar{q_j} - (\bar{q_j} + \delta) p_i \big)
	&< \sum_{k=T}^N g(k)  \big(\delta p_i \big)
\end{align}

Let $g(k) = \gamma^k$, with $\gamma \in [0,1]$. In this case, any
choice of $T$ can be made equivalent to $T=1$ by dividing everything
with $\sum_{k=0}^{T-1} \gamma^k$. We explore two cases: $\gamma < 1, ~ N
\rightarrow \infty$ and $\gamma = 1, ~ N <
\infty$. In the first case, which corresponds to infinite horizon
exponentially discounted reward maximisation problems, we obtain the
following:
\begin{align}
\bar{q_j} - (\bar{q_j} + \delta) p_i &< \sum_{k=1}^\infty \gamma^k \delta p_i\\
%%\bar{q_j} - (\bar{q_j} + \delta) p_i &< \frac{\gamma}{1-\gamma} \delta p_i\\
%%\bar{q_j} - (\bar{q_j} + \delta) p_i &< \gamma (\delta p_i +  \bar{q_j} - (\bar{q_j} + \delta)
%%p_i)\\
\frac{\bar{q_j} - (\bar{q_j} + \delta) p_i}{(1-p_i)\bar{q_j}} &< \gamma.
\label{eq:gamma_condition}
\end{align}
It possible to simplify this expression
considerably. When $ P(q_i \geq \bar{q_j} + \delta) = 1/2$, it follows from
\eqref{eq:gamma_condition} that
\begin{equation}
\label{eq:gamma_median_condition}
\gamma > \frac{\bar{q_j} - (\bar{q_j} + \delta) /2 }{\bar{q_j} / 2} = \frac{\bar{q_j}-\delta}{\bar{q_j}}.
\end{equation}
Thus,  for infinite horizon discounted reward
maximisation problems, when it is known that the all expected rewards
are non-negative, all we need to do is find $\delta$ such that $P(q_i
\geq q_j + \delta) = 1/2$. Then \eqref{eq:gamma_median_condition} can
be used to make a decision on whether it is worthwhile to perform
exploration.  Although it might seem strange the $q_i$ is omitted from
this expression, its value is implicitly expressed through the value
of $\delta$.

In the second case, finite horizon cumulative reward maximisation
problems, exploration should be performed when the following condition
is satisfied:
\begin{align}
    \label{eq:finite_horizon_condition}
    N \delta p_i &>    \bar{q_j} - (\bar{q_j} + \delta) p_i 
\end{align}
Here the decision making function is of a different nature, since it
depends on both estimates. However, in both cases, the longer the
effective horizon becomes and the larger the uncertainty is, the more
the bias towards exploration is increased.  We furthermore note that
in the finite horizon case, the backward induction procedure can be
used to make optimal decisions (see
\cite{Degroot:OptimalStatisticalDecisions} Sec.~12.4).

\subsection{Solutions for Specific Distributions}

If we have a specific form for the distribution $P(q_i > q_j +
\delta)$ it may be possible to obtain analytical solutions.  To see
how this can be achieved, consider that from
\eqref{eq:gamma_condition}, we have:
\begin{align}
\gamma \bar{q_j} &> \bar{q_j} - \delta \frac{p_i}{1-p_i} \nonumber \\
%%0 &< \delta \frac{p_i}{1-p_i} - (1-\gamma) \bar{q_j} \nonumber \\
0 &< \delta \frac{P(q_i>q_j+\delta)}{1-P(q_j>q_j + \delta)} - (1-\gamma)
\bar{q_j},
\label{eq:gamma_dist_condition}
\end{align}
recalling that all mean rewards are non-negative.  

If this condition is satisfied for some $\delta$ then exploration must
be performed.  We observe that if the first term is maximised for some
$\delta^*$ for which the inequality is not satisfied, then there is no
$\delta \neq \delta^*$ that can satisfy it.  Thus, we can attempt to
examine some distributions for which this $\delta^*$ can be
determined.  We shall restrict ourselves to distributions that are
bounded below, due to Assumption \ref{ass:bounded_rewards}.

\subsection{Solutions for the Exponential Distribution}
One such distribution is the exponential distribution, defined as
\[
P(X>\delta) =
\int_\delta^\infty \beta e^{-\beta(x-\mu)} dx
 = e^{-\beta(\delta-\mu)}
\]
if $\delta > \mu$, 1 otherwise.
We may plug this into \eqref{eq:gamma_dist_condition} as follows
\[
f(\delta) = \delta \frac{P(q_i>q_j+\delta)}{1-P(q_i>q_j + \delta)}
= \delta
\frac{e^{-\beta_i(\mu_j + \delta-\mu_i)}}
  {1-e^{-\beta_i(\mu_j + \delta-\mu_i)}}
=
\frac{\delta}
  {e^{\beta_i(\mu_j + \delta-\mu_i)} - 1}
\]
Now we should attempt to find $\delta^*=\argmax_\delta f(\delta)$.  We
begin by taking the derivative with respect to $\delta$.
Set $g(\delta)=e^{h(\delta)} - 1$, $h(\delta)=\beta_i(\bar{q_j} + \delta-\mu_i)$
\begin{align*}
  \nabla f(\delta)
      & =  \frac{g(\delta) - \delta \nabla g(\delta)}
              {g(\delta)^2}
      =  \frac{g(\delta) - \delta \beta_i \nabla_h g(\delta)}
              {g(\delta)^2}
     =  \frac{e^{h(\delta)} (1 - \delta \beta_i) -1}
            {(e^{h(\delta)} - 1)^2}
\end{align*}
%\begin{align*}
%  \nabla f(\delta)
%      & =  \frac{g(\delta) - \delta \nabla g(\delta)}
%              {(e^{h(\delta)} - 1)^2}
%    & =  \nabla \frac{\delta} {g(\delta)}
%      =  \frac{g(\delta) - \delta \nabla g(\delta)}
%              {g(\delta)^2}\\
%    & =  \frac{g(\delta) - \delta \nabla h \nabla_h g(\delta)}
%              {g(\delta)^2}
%      =  \frac{g(\delta) - \delta \beta_i \nabla_h g(\delta)}
%              {g(\delta)^2}\\
%    &  =  \frac{e^{h(\delta)} - 1 - \delta \beta_i e^{h(\delta)}}
%              {e^{2h(\delta)} - 2e^{h(\delta)} + 1}
%     =  \frac{e^{h(\delta)} (1 - \delta \beta_i) -1}
%              {(e^{h(\delta)} - 1)^2}
%\end{align*}
Necessary and sufficient conditions for some point $\delta^*$ to be a
local maximum for a continuous differentiable function $f(\delta)$ are
that $\nabla_\delta f(\delta^*)=0$ and $\nabla_\delta^2 f(\delta^*) <
0$. 
The necessary condition for $\delta$ results in
\begin{align}
\label{eq:delta_for_exp}
e^{\beta_i(q_k + \delta - \mu_i)}(1 - \delta \beta_i)  = 1.
\end{align}
Unfortunately \eqref{eq:delta_for_exp} has no closed form solution,
but it is related to the Lambert W function for which iterative
solutions do exist~\cite{acm:corless:1996}.  The found solution can
then be plugged into
\eqref{eq:gamma_dist_condition} to see whether the conditions for
exploration are satisfied.

\section{Extension to the General Case}
\label{sec:optimal_reinforcement_learning}
In the general reinforcement learning setting, the reward distribution
does not only depend on the action taken but additionally on a state
variable.  The state transition distribution is conditioned on actions
and has the Markov property.  Each particular task within this
framework can be summarised as a Markov decision process:
\begin{definition}[Markov decision process]
A Markov decision process is defined by a set of states $\CS$, a set
of actions $\CA$, a transition distribution $\FT(s', s, a) =
P(s_{t+1}'|s_t = s, a_t = a)$ and a reward distribution $\FR(s', s, a)
= p(r_{t+1}|s_{t+1}=s', s_t = s, a_t = a)$.
\end{definition}
The simplest way to extend the bandit case to the more general one of
MDPs is to find conditions under which the latter reduces to the
former.  This can be done for example by considering choices not
between simple actions but between {\em temporally extended actions},
which we will refer to as {\em options} following
\cite{sutton99between}.  We shall only need a simplified version of
this framework, where each possible {\em option} $x$ corresponds to
some policy $\pi^x : \CS \times \CA \to [0,1]$.  This is sufficient
for sketching the conditions under which the equivalence arises.

In particular, we examine the case where we have two options. The
first option is to always select actions according to some exploratory
principle, such picking them from a uniform distribution. The second
is to always select actions greedily, i.e. by picking the action with
the highest expected return.

We assume that each option will last for time $T$. One further
necessary component for this framework is the notion of mixing time
\begin{definition}[Exploration mixing time]
	We define the exploration mixing time for a particular MDP
	$\CM$ and a policy $\pi$ $T_\epsilon(\CM, \pi)$ as the
	expected number of time steps after which the state
	distribution is close to the stationary state distribution of
	$\pi$ after we have taken an exploratory action $i$ at time
	step $t$, i.e. the expected number of steps $T$ such that the
	following condition holds:
\[
	\frac{1}{\|\BS\|} \sum_s \|P(s_{t+T}=s|s_{t},\pi) -
        P(s_{t+T}=s|a_t=i, s_t,\pi)\| < \epsilon
\]	
\end{definition}
It is of course necessary for the MDP to be ergodic for this to be
finite.  If we only consider switching between options at time periods
greater than $T_\epsilon(\CM, \pi)$, then the option framework's
roughly corresponds to the bandit framework, and $T_\epsilon$ in the
former to $T$ in the latter.  This means that whenever we take an
exploratory action $i$ (one that does not correspond to the action
that would have been selected by the greedy policy $\pi$), the
distribution of states would remain to be significantly different from
that under $\pi$ for $T_\epsilon(\CM, \pi)$ time steps.  Thus we could
consider the exploration to be taking place during all of
$T_\epsilon$, after which we would be free to continue exploration or
not.  Although there is no direct correspondence between the two
cases, this limited equivalence could be sufficient for motivating the
use of similar techniques for determining the optimal exploration
exploitation threshold in full MDPs.

\section{Optimistic Evaluation}
\label{sec:optimistic_evaluation}

In order to utilise Lemma~\ref{lem:exploration_bound} in a practical
setting we must define $T$ in some sense. 
% This can be done by
%calculating whether $P(q_i<q_j)$ is smaller than some arbitrarily
%small probability threshold $\varepsilon$, for example by
%\begin{inparaenum}[(a)] \item utilising Chebyshev bounds on sums, \item using a
%Bayesian framework and performing analytical or approximate
%integration or \item implementing the distribution as a population,
%(again in a Bayesian framework)
%\end{inparaenum}.
The simplest solution is to set $T=1$, which results in an optimistic
estimate for exploratory actions as will be shown below.  By
rearranging
\eqref{eq:exploration_bound} we have
\begin{align}
U(i, j, T, \delta, b)
%%    &= \sum_{k=0}^N g(k) (\bar{q_j} + \delta) p_i
%%      + (1-p_i) \left(\sum_{k=0}^{T-1} g(k) b + \sum_{k=T}^N g(k)
%%\bar{q_j}\right)\\
    &= \sum_{k=0}^N g(k) \bar{q_j}  + \sum_{k=0}^N g(k) \delta p_i 
      + (1-p_i) \left(\sum_{k=0}^{T-1} g(k) (b  - \bar{q_j}))\right)
\end{align}
from which it is evident, since $q_j \geq b$ and $g(k) \geq 0$, that
$U(i,j,T_1,\delta,b) \geq U(i,j,T_2,\delta,b)$ when $T_1 < T_2$, thus
$U(i,j,1,\delta, b) \geq U(i,j,T,\delta, b)$ for any $T \geq 1$. 
This can now be used to obtain Alg. \ref{al:optimistic} for
optimistic exploration.

Nevertheless, testing for the existence of a suitable $\delta$ can be
costly since, barring an analytic procedure it requires an exhaustive
search.  On the other hand, it may be possible to achieve a similar
result through sampling for different values of $\delta$.  Herein, the
following sampling method is considered: Firstly, we determine the
action $j$ with the greatest $\bar{q_j}$. Then, for each action $i$ we
take a sample $x$ from the distribution $p(q_i)$ and set $\delta = x -
\bar{q_j}$.  This is quite an arbitrary sampling method, but we may
expect to obtain a $\delta > 0$ with high probability if $i$ has a
high probability to be significantly better than $j$.  This method is
summarised in Alg. \ref{al:optimistic-stochastic}.
\begin{algorithm}
    \caption{Optimistic exploration}
    \label{al:optimistic}
    \begin{algorithmic}
        \IF{$\exists ~\delta ~ : ~U(i,j,1,\delta, b) > \sum_{k=0}^N g(k) \bar{q_j}$}
        \STATE $a \Leftarrow i$
        \ELSE
        \STATE $a \Leftarrow j$
        \ENDIF
    \end{algorithmic}
\end{algorithm}
\begin{algorithm}
    \caption{Optimistic stochastic exploration}
    \label{al:optimistic-stochastic}
    \begin{algorithmic}
        \STATE $j \Leftarrow \argmax_i \bar{q_i}$.
        \STATE $u_j = \sum_{k=0}^N g(k) \bar{q_j}$.
        \FORALL{$i \neq j$}
            \STATE $\delta \Leftarrow x - \bar{q_j}, \quad x \sim p(q_i)$
            \STATE $u_i \Leftarrow  U(i,j,1,\delta, b)$
        \ENDFOR
        \STATE  $a \Leftarrow \argmax_i u_i$
    \end{algorithmic}
\end{algorithm}
\begin{algorithm}
    \caption{Thompson sampling}
    \label{al:Thompson sampling}
    \begin{algorithmic}
        \STATE $a \Leftarrow i$ with probability
            $P(a=i) = P(q_i > q_j) ~ \forall j \neq i$
    \end{algorithmic}
\end{algorithm}

An alternative exploration method is given by
Alg. \ref{al:Thompson sampling}, first introduced in\cite{thompson1933lou}, which samples each action with
probability equal to the probability that its expected reward is the
highest. It can perhaps be viewed as an approximation to
Alg. \ref{al:optimistic-stochastic} when $\gamma \rightarrow 1$ and
has the advantage that it is extremely simple. However, both of these algorithms require us to be able to sample from the posterior distribution, something that may not always be feasible. For that reason, in the next section we introduce the idea of online bootstrap for reinforcement learning.

%For $g(k) = \gamma^k$, we obtain the following simple expression:
%\begin{equation}
%    U(i,j,1,\delta,0) - \sum_{k=0}^N \gamma^k \bar{q_j}
%    = \frac{1}{1-\gamma} \delta p_i
%      -  (1 - p_i) \bar{q_j}.
%\end{equation}
\section{Online Bootstrapping for reinforcement learning}
\label{sec:online-bootstr-reinf}

The algorithms we consider require sampling from a posterior distribution. However, there are two main problems associated with this requirement. Firstly, calculating posterior distributions and sampling from them may be computationally demanding. Secondly, the Bayesian framework requires specifying some prior distribution, and it may not always be clear what this should be. Even when we simply wish to estimate reward distributions, for example, without specific domain knowledge it is not easy to make a choice between standard conjugate prior distributions such as the Beta and Normal-Gamma. While this latter problem can be alleviated by using a hierarchical prior, or a non-parametric approach, this paper proposes a simpler solution:  using online bootstrap samples to approximate posteriors.

For bandit problems in particular, we maintain a population of $K$ point estimates for the expected reward of each arm $i$, so that at time $t$ the $k$-th estimate is $q_i^k(t)$. The initial values $q_j^i(0)$ are drawn i.i.d. from an appropriate prior distribution. Then, at each step  $t$, we draw $k_t$ uniformly from $\{1, K\}$ and then use $q_i^{k_t}(t)$ as our mean reward for both optimistic stochastic exploration and Thompson sampling. Finally, we perform online bootstrapping using
\begin{equation}
  \label{eq:bootstrap}
  q_{a_t}^{k_t}(t) = (1 - \alpha_t^k)  q_{a_t}^{k_t}(t-1)  + (1 - \alpha_t^k)  r_t.
\end{equation}
If $\alpha_t^k = \alpha_t$ for all $k$, then the only diversity in the population estimate will be due to the initial values. Randomising $\alpha_t^k$ in such a way that the same conditions for convergence hold will ensure that population diversity will be maintained in a way that reflects uncertainty due to the data.\footnote{One such example, not used in this paper, is $\alpha_t^k = \alpha_t \nu_t^k$, where $\nu_t^k$ is sampled from an exponential distribution. The simplest approach to implement, however, is to sample $\nu_t^k$ from a Bernoulli distribution so that only some population members are updated.}
Although we only applied this algorithm for bandit problems, it is also possible to apply it to the transition kernel or value function for the full reinforcement learning problem. However, care should be taken to avoid problems with correlations in the bootstrap samples. It is probably to difficult to apply this technique to non-episodic problems in particular, without domain-specific knowledge.

\section{Experiments}
\label{sec:optimal_bandit_experiments}
A small experiment was performed on a $n$-armed bandit problem with
rewards $r_t \in \{0, 1\}$ drawn from a Bernoulli distribution.
Alg. \ref{al:optimistic-stochastic} was used with $g(k) = \gamma^k$
and $b=0$, which is in agreement with the distribution.  This was
compared with Alg. \ref{al:Thompson sampling}, which can be perhaps
viewed as a crude approximation to Alg. \ref{al:optimistic-stochastic}
when $\gamma \rightarrow 1$.  The performance of $\epsilon$-greedy
action selection with $\epsilon = 0.01$ was evaluated for reference.

The $\epsilon$-greedy algorithm used point estimates for $\bar{q}_i$,
which were updated with gradient descent with a step size of $\alpha =
0.01$, such that for each action-reward observation tuple
$(a_t=i,r_{t+1})$, $\bar{q}_i \Leftarrow \alpha (r_{t+1} -
\bar{q}_i)$, with initial estimates being uniformly distributed in
$[0,1]$.  In the other two cases, the complete distribution of $q_i$
was maintained via a population $\{p_i^k\}_{k=0}^K$ of point estimates,
with $K=16$.  Each point estimate in the population was maintained in
the same manner as the single point estimates in the $\epsilon$-greedy
approach.  Sampling actions was performed by sampling uniformly from
the members of the population for each action.

\begin{figure}[htb]
  \begin{center}
    \subfigure{\epsfig{figure=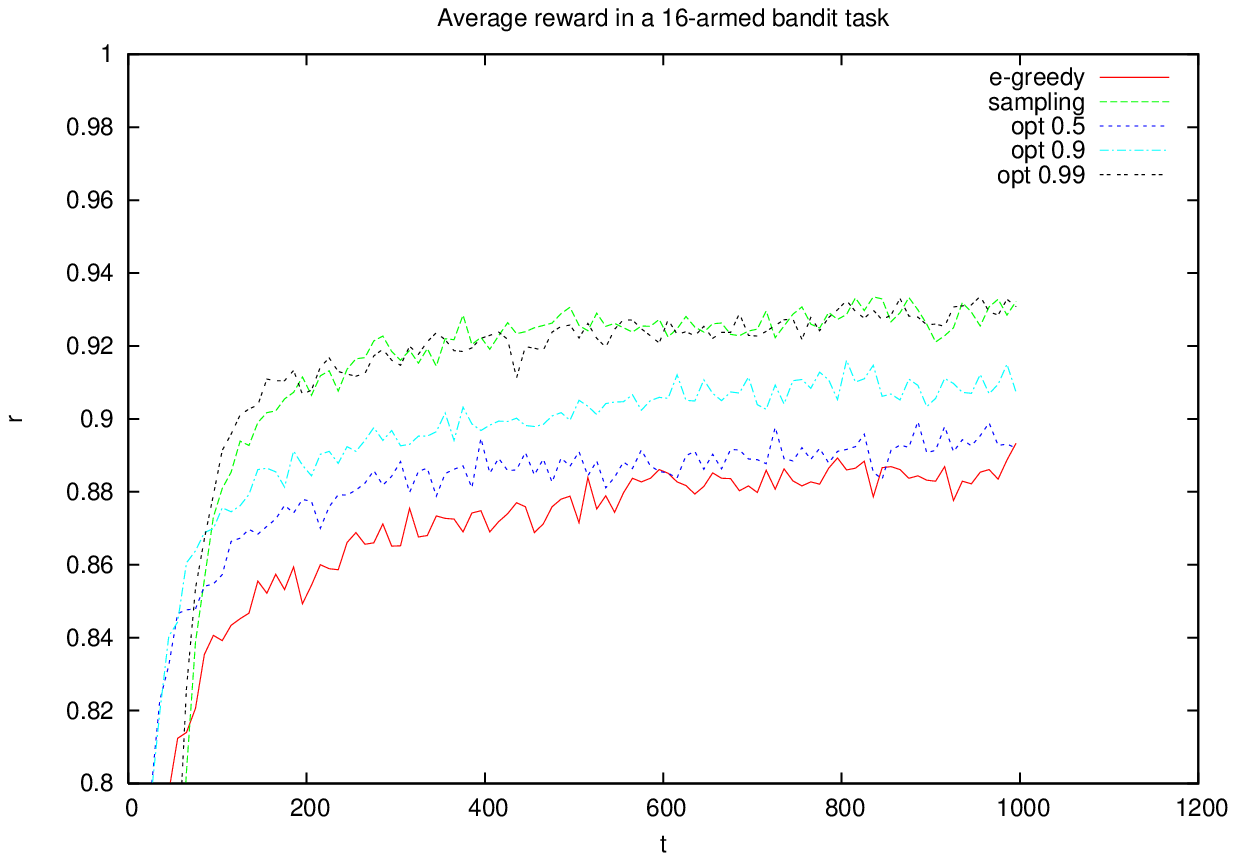,width=.49\linewidth}}
    \subfigure{\epsfig{figure=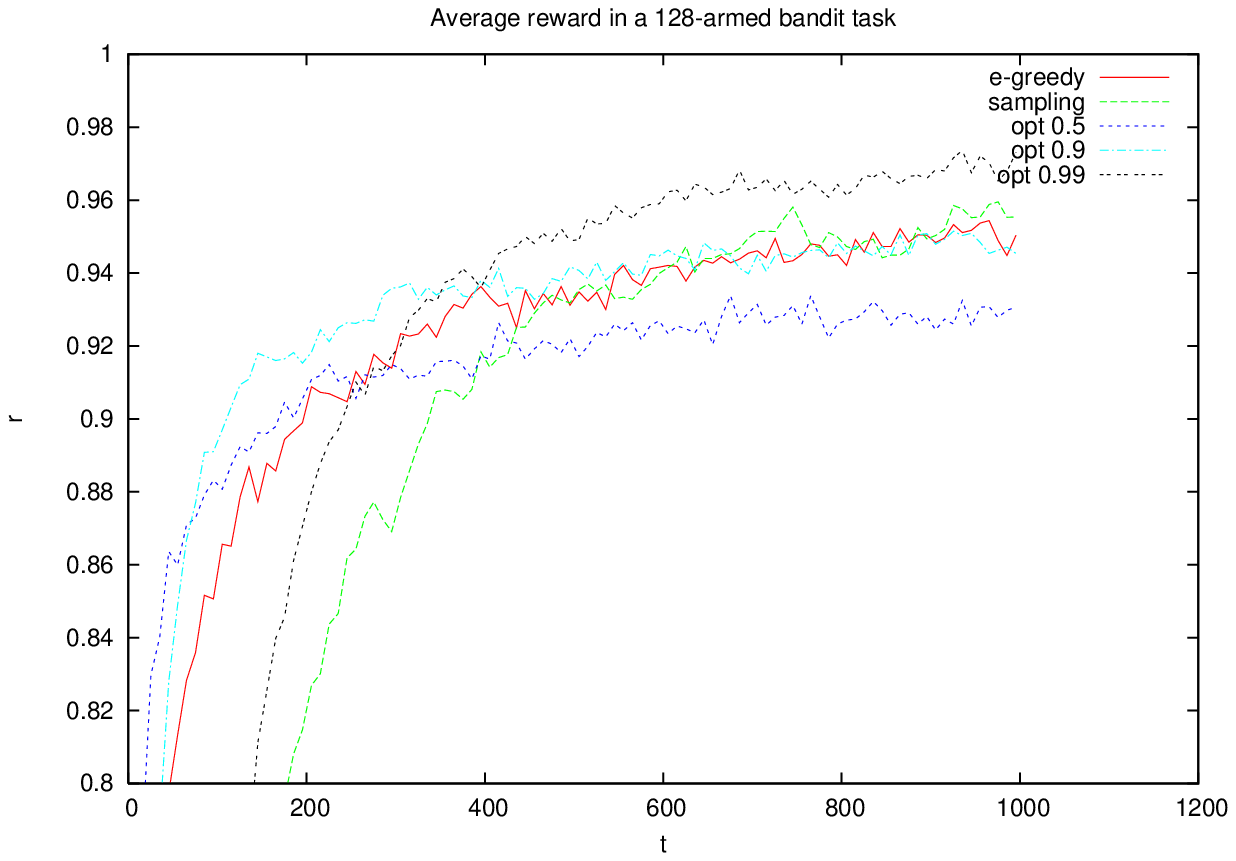,width=.49\linewidth}}
	\caption[Average reward in a bandit task]{Average reward in an
    	multi-armed bandit task averaged over 1,000 experiments,
        smoothed with a moving average over 10 time-steps.  Results
	    are shown for $\epsilon$-greedy ({\bf e-greedy}),
    	Thompson sampling ({\bf sampling}) and  Alg.
        \ref{al:optimistic-stochastic} ({\bf opt}) with
        $\gamma \in \{0.5, 0.9, 0.99\}$. }
    \end{center}
    \label{fig:bandit_task_averages}
\end{figure}

The results for two different bandit tasks, one with 16 and the other
with 128 arms, averaged over 1,000 runs, are summarised in
Fig.~\ref{fig:bandit_task_averages}.  For each run, the expected
reward of each bandit was sampled uniformly from $[0,1]$.
As can be seen from the figure, the $\epsilon$-greedy approach
performs relatively well when used with reasonable first initial
estimates.  The Thompson sampling approach, while having apprximately
the same complexity, appears to perform better asymptotically.  More
importantly, Alg.~\ref{al:optimistic-stochastic} exhibits better
long-term versus short-term performance when the effective reward
horizon is increased as $\gamma \rightarrow 1$, which means that the algorithm does manage to take into account the horizon when deciding how to explore.

\begin{figure}[htb]
  \begin{center}
    \epsfig{figure=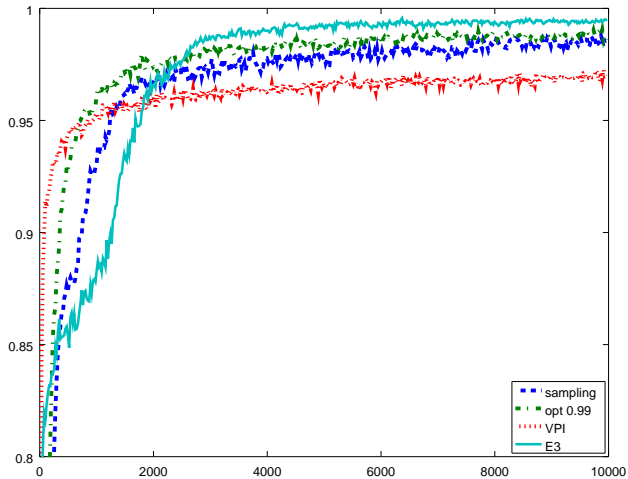,width=.80\linewidth}
	\caption[Average reward in a bandit task]{Average reward in an
    	multi-armed bandit task averaged over 1,000 experiments,
        smoothed with a moving average over 10 time-steps.  Results
	    are shown for Thompson sampling ({\bf sampling}) and  Alg.
        \ref{al:optimistic-stochastic} ({\bf opt}), VPI {\bf VPI} and $E^3$ ({\bf E3}).}
    \end{center}
    \label{fig:bandit_task_optimal}
\end{figure}

Finally, we show results for a bandit task with 256 arms, averaged over 1,000 runs, in comparison with the $E^3$ and VPI algorithms in
Fig.~\ref{fig:bandit_task_optimal}.  
There, we can see that the optimistic sampling approach clearly dominates Thompson sampling, at least under the bootstrap distribution used. Only $E^3$ has better asymptotic performance in this task

%In order to utilise a full Bayesian framework, we consider the case
%when the rewards are Bernoulli random variables with probability of
%success $q$, such that $q_i = E[r_{t+1}|a_t=i]$, $r_t \in \{0,1\}$.
%Assume that the prior distribution of $q$ is a beta distribution
%\[
%    p(q) = \frac{q^{a-1}(1-q)^{b-1}}{B(a,b)}.
%\]
%with $a, b > 0$ and $B(a,b) = \int_0^1 x^{a-1}(1-x)^{b-1} dx$.
%Since these are conjugate (see
%\cite{Degroot:OptimalStatisticalDecisions} Section 9.2), after
%observing $s$ successes and $f$ failures, the posterior becomes
%the beta distribution
%\[
%    p(q) = \frac{q^{a+s-1}(1-q)^{b+f-1}}{B(a+s,b+f)}.
%\]
%Let's apply this to 

\section{Discussion and Conclusion}
This paper has presented a formulation of an optimal
exploration-exploitation threshold for in a $n$-armed bandit task,
which links the need for exploration to the effective reward horizon
and model uncertainty.  Additionally, a practical algorithm, based on
an optimistic bound on the value of exploration, is introduced.
Experimental results show that this algorithm exhibits the expected
long-term versus short-term performance trade-off when the effective
reward horizon is increased.

While the above formulation fits well within a reinforcement learning
framework, other useful formulations may exist.  In {\em budgeted
learning}, any exploratory action results in a fixed cost.  Such a
formulation is used in \cite{COLT:Madani:Budgeted:2004} for the bandit
problem (see also \cite{UAI:Madani:ActiveModelSelection:2004} for the
active learning case).  Then the problem essentially becomes that of
how to best sample from actions in the next $T$ moves such that the
expected return of the optimal policy {\em after} $T$ moves is
maximised and corresponds to $g(k)=0 ~ \forall k < T$ in the framework
presented in this paper.  A further alternative, described in
\cite{jmlr:Even-Dar:ActionElimination}, is to stop exploring those
parts of the state-action space which lead to sub-optimal returns with
high probability.

When a distribution or a confidence interval is available for expected
returns, it is common to use the optimistic side of the confidence
interval for action selection \cite{pascal:auer:2005}.  This practice
can be partially justified through the framework presented herein, or
alternatively, through considering maximising the expected information
to be gained by exploration, as proposed by
\cite{AOS:Berndardo:UtilityAndInformation:1979}.  In a similar manner,
other methods which represent uncertainty as a simple additive factor
to the normal expected reward estimates, acquire further meaning when
viewed through a statistical decision making framework.  For example
the Dyna-Q+ algorithm (see \cite{Sutton+Barto:1998} chap. 9) includes
a slowly increasing {\em exploration bonus} for state-action pairs
which have not been recently explored.  From a statistical viewpoint,
the exploration bonus corresponds to a model of a non-stationary
world, where uncertainty about past experiences increases with elapsed
time elapsed.

In general, the conditions defined in Sec. \ref{sec:optimal_bandit}
require maintaining some type of belief, int the form of a distribution, over the expected
return of actions. A natural choice for this
would be to use a fully analytical Bayesian framework. 
Unfortunately this makes it more difficult to calculate $P(q_i>d)$, so
it might be better to consider simple numerical approaches from the
outset, such as the bootstrap ones proposed in this paper.  We have previously considered some simple such estimates in
\cite{dimitrakakis:rr05-29}, where we relied on estimating the
gradient of the expected return with respect to the parameters.  The
estimated gradient was then used as a measure of uncertainty.  Further
results on the use of ensemble estimates for reinforcement learning, and  comparisons between various forms of bootstrapping with particle filters and analytical Bayesian estimates can be found in a longer version of this paper in~\cite{christos-dimitrakakis:phd-thesis:2006}.

\bibliographystyle{splncs}
\bibliography{../../bib/mine,../../bib/nc,../../bib/misc,../../bib/nn,../../bib/npl,../../bib/HKP,../../bib/nn-books,../../bib/nips-7,../../bib/mach,../../bib/PRNN}

\begin{thebibliography}{10}

\bibitem{Wald:SequentialAnalysis}
Wald, A.:
\newblock Sequential Analysis.
\newblock John Wiley \& Sons (1947) Republished by Dover in 2004.

\bibitem{Degroot:OptimalStatisticalDecisions}
De{G}root, M.H.:
\newblock Optimal Statistical Decisions.
\newblock John Wiley \& Sons (1970)

\bibitem{Bellman:1956}
Bellman, R.E.:
\newblock A problem in the sequential design of experiments.
\newblock Sankhya \textbf{16} (1957)  221--229

\bibitem{jmlr:Mannor:SampleComplexity}
Mannor, S., Tsitsiklis, J.N.:
\newblock The sample complexity of exploration in the multi-armed bandit
  problem.
\newblock Journal of Machine Learning Research \textbf{5} (2004)  623--648

\bibitem{dearden98bayesian}
Dearden, R., Friedman, N., Russell, S.J.:
\newblock {Bayesian} {Q}-learning.
\newblock In: {AAAI}/{IAAI}. (1998)  761--768

\bibitem{jmlr:Even-Dar:ActionElimination}
Even-Dar, E., Mannor, S., Mansour, Y.:
\newblock Action elimination and stopping conditions for the multi-armed and
  reinforcement learning problems.
\newblock Journal of Machine Learning Research (2006)  1079--1105

\bibitem{Sutton+Barto:1998}
Sutton, R.S., Barto, A.G.:
\newblock Reinforcement Learning: An Introduction.
\newblock MIT Press (1998)

\bibitem{acm:corless:1996}
Corless, R.M., Gonnet, G.H., Hare, D.E.G., Jeffrey, D.J., Knuth, D.E.:
\newblock On the lambert {W} function.
\newblock Advances in Computational Mathematics \textbf{5} (1996)  329--359

\bibitem{sutton99between}
Sutton, R.S., Precup, D., Singh, S.P.:
\newblock Between {MDPs} and semi-{MDPs}: A framework for temporal abstraction
  in reinforcement learning.
\newblock Artificial Intelligence \textbf{112}(1-2) (1999)  181--211

\bibitem{thompson1933lou}
Thompson, W.:
\newblock {On the Likelihood that One Unknown Probability Exceeds Another in
  View of the Evidence of two Samples}.
\newblock Biometrika \textbf{25}(3-4) (1933)  285--294

\bibitem{COLT:Madani:Budgeted:2004}
Madani, O., Lizotte, D.J., Greiner, R.:
\newblock The budgeted multi-armed bandit problem.
\newblock In: Learning Theory: 17th Annual Conference on earning Theory, COLT
  2004. Volume 3120 of Lecture Notes in Computer Science., Springer (2004)
  643--645

\bibitem{UAI:Madani:ActiveModelSelection:2004}
Madani, O., Lizotte, D.J., Greiner, R.:
\newblock Active model selection.
\newblock In: Proceedings of the 20th Conference on Uncertainty in Artificial
  Intelligence, Banff, Canada, AUAI Press, Arlington, Virginia (2004)  357--365

\bibitem{pascal:auer:2005}
Auer, P.:
\newblock Models for trading exploration and exploitation using upper
  confidence bounds.
\newblock In: PASCAL workshop on principled methods of trading exploration and
  exploitation, PASCAL Network (2005)

\bibitem{AOS:Berndardo:UtilityAndInformation:1979}
Bernardo, J.M.:
\newblock Expected information as expected utility.
\newblock In: The Annals of Statistics. Volume~7., Institute of Mathematical
  Statistics (1979)  686--690

\bibitem{dimitrakakis:rr05-29}
Dimitrakakis, C., Bengio, S.:
\newblock Gradient estimates of return.
\newblock IDIAP-RR 05-29, IDIAP (2005)

\bibitem{christos-dimitrakakis:phd-thesis:2006}
Dimitrakakis, C.:
\newblock Ensembles for Sequence Learning.
\newblock PhD thesis, {\'E}cole Polytechnique F{\'e}d{\'e}rale de Lausanne
  (2006)

\end{thebibliography}

\end{document}